\documentclass{article} 
\PassOptionsToPackage{numbers, compress}{natbib}
\usepackage[preprint]{style/style}
\usepackage{times}

\usepackage{amsmath,amsfonts,bm}

\def\eqref#1{equation~\ref{#1}}

\def\1{\bm{1}}

\def\mW{{\bm{W}}}

\DeclareMathAlphabet{\mathsfit}{\encodingdefault}{\sfdefault}{m}{sl}
\SetMathAlphabet{\mathsfit}{bold}{\encodingdefault}{\sfdefault}{bx}{n}

\def\gD{{\mathcal{D}}}

\def\gG{{\mathcal{G}}}

\def\gO{{\mathcal{O}}}

\def\gX{{\mathcal{X}}}

\newcommand{\E}{\mathop{{}\mathbb{E}}}

\newcommand{\R}{\mathbb{R}}

\DeclareMathOperator*{\argmin}{arg\,min}

\usepackage{comment}
\usepackage{hyperref}
\usepackage{url}
\usepackage{physics}
\usepackage{amsthm}
\usepackage{amssymb}
\usepackage{amsmath}
\usepackage{graphicx}
\usepackage{algpseudocode}
\usepackage{algorithm}
\usepackage{caption}
\usepackage{subcaption}
\usepackage{enumitem}
\usepackage{bbm}
\usepackage{wrapfig}

\newcommand{\bx}{\mathbf{x}}
\newcommand{\X}{\mathbf{x}}

\newcommand{\A}{\mathbf{A}}

\newcommand{\G}{\mathcal{G}}

\newcommand{\btheta}{\boldsymbol{\theta}}

\newcommand{\kl}{\operatorname{D_{KL}}}

\newcommand{\appropto}{\mathrel{\vcenter{
  \offinterlineskip\halign{\hfil$##$\cr
    \propto\cr\noalign{\kern2pt}\sim\cr\noalign{\kern-2pt}}}}}
    
    \usepackage[dvipsnames]{xcolor}

\theoremstyle{definition}

\theoremstyle{definition}

\theoremstyle{plain}
\newtheorem{proposition}{Proposition}
\theoremstyle{plain}

\theoremstyle{plain}

\theoremstyle{remark}

\title{Variational Causal Networks: Approximate Bayesian Inference over Causal Structures}
\author{%
    \parbox{\linewidth}{
        \centering 
        Yashas Annadani $^{1}$\thanks{Work done during an internship at Mila.}, 
        Jonas Rothfuss $^1$, 
        Alexandre Lacoste $^3$, 
        Nino Scherrer $^1$, \\
        Anirudh Goyal $^2$, 
        Yoshua Bengio $^2$,
        Stefan Bauer $^{4,5}$
    }\\
    \parbox{\linewidth}{
        \centering 
        $^1$ ETH Zurich, 
        $^2$ Mila, Université de Montréal,
        $^3$ ElementAI/ ServiceNow,
        $^4$ MPI for Intelligent Systems
        $^5$ CIFAR Azrieli Global Scholar
    }}
\begin{document}

\maketitle

\begin{abstract}
\looseness -1 Learning the causal structure that underlies  data is a crucial step towards robust real-world decision making. The majority of existing work in causal inference focuses on determining a single \emph{directed acyclic graph (DAG)} or a Markov equivalence class thereof. However, a crucial aspect to acting intelligently upon the knowledge about causal structure which has been inferred from finite data demands reasoning about its uncertainty. For instance, planning \emph{interventions} to find out more about the causal mechanisms that govern our data requires quantifying  \emph{epistemic uncertainty} over DAGs. While Bayesian causal inference allows to do so, the posterior over DAGs becomes intractable even for a small number of variables. Aiming to overcome this issue, we propose a form of variational inference over the graphs of Structural Causal Models (SCMs). To this end, we introduce a parametric variational family modelled by an autoregressive distribution over the space of discrete DAGs. Its number of parameters does not grow exponentially with the number of variables and can be tractably learned by maximising an Evidence Lower Bound (ELBO). In our experiments, we demonstrate that the proposed variational posterior is able to provide a good approximation of the true posterior.

\end{abstract}

\begin{wrapfigure}{r}{0.45\textwidth}
    \centering
    \vspace{-2mm}
    \includegraphics[height=5cm]{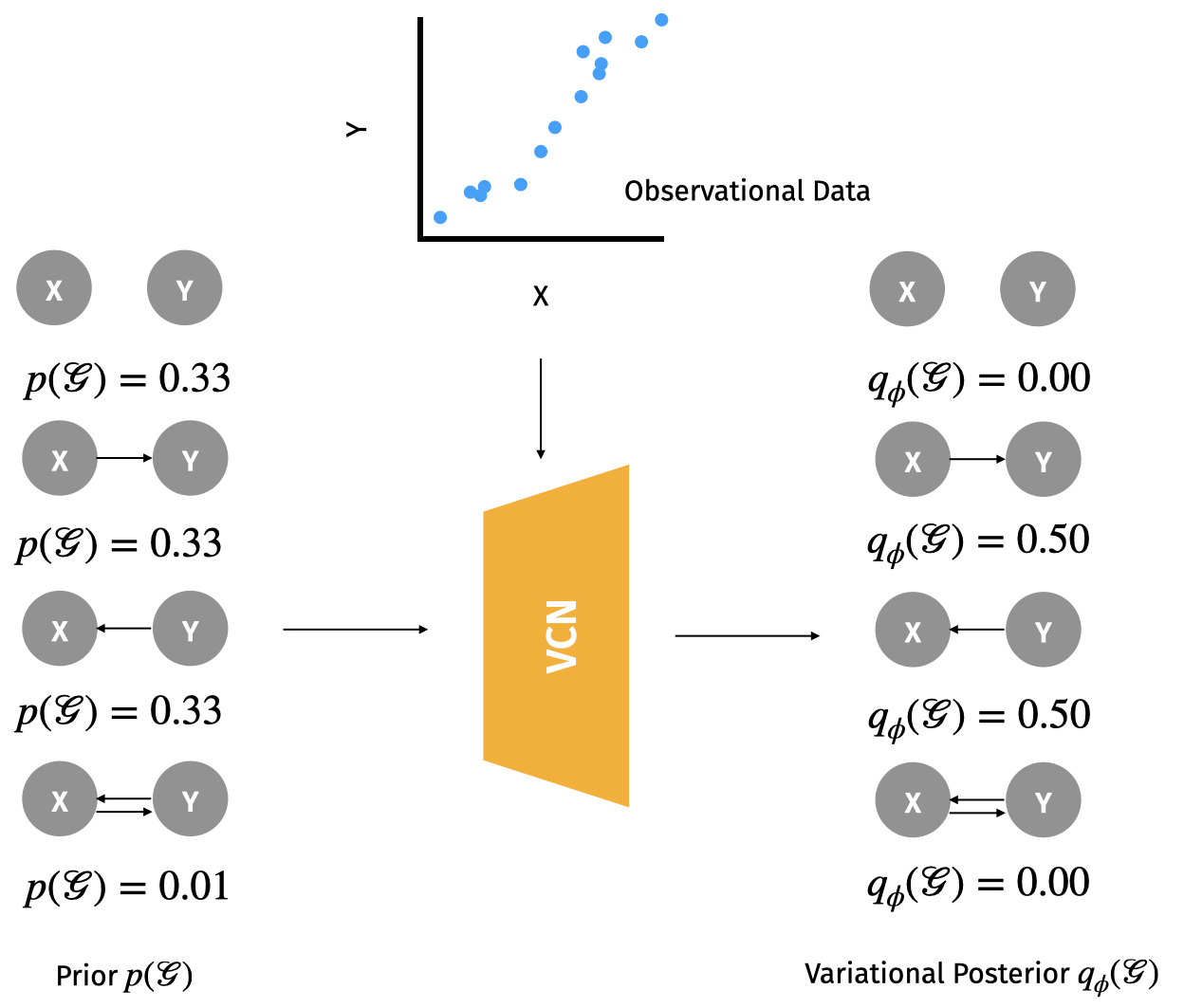}
    \caption{Schematic Diagram of Variational Inference on SCMs with VCN.} 
    \label{fig:vi}
\end{wrapfigure}

Moving from learning correlation and association in data to causation is a critical step towards increased robustness, interpretability and real-world decision-making \citep{pearl2009causality,spirtes2000causation}. Doing so entails learning the causal structure underlying the data generating process. Causal inference is concerned with determining the causal structure of a set of random variables from data, commonly represented as a directed acycilc graph (DAG) \citep{peters2017elements}.  While Structural Causal Models (SCMs) provide a generative model over the data, they are hard to learn from data due to the non-identifiability of the causal models without interventional data \citep{eberhardt2007interventions} and the combinatorial nature of the space of DAGs \citep{heinze2018causal}.  Even with infinite amount of data, recovering the causal structure is intrinsically hard since a DAG is only identifiable up to its \emph{Markov equivalence class} (MEC) and the space of possible DAGs grows super-exponentially with the number of variables. While the majority of work on causal inference \citep{chickering2002optimal, buhlmann2014cam, zheng2018dags} deals with getting a single underlying causal structure without a probabilistic treatment, quantifying the epistemic uncertainty in case of non-identifiability is crucial and is not possible in these approaches.

In this work, we take a Bayesian approach to causal structure learning. Given {\em only finite observational data}, a Bayesian approach allows us to quantify the uncertainty in the causal structure of the data generating process, even before performing interventions. Having such a framework over causal structures can help further downstream tasks on graph learning and causal inference. For example, we can leverage the model's uncertainty to select informative interventions and discover the full graph with minimal amount of interventions~\citep{agrawal2019abcd}. Also, for estimating the causal effect of certain variables, it is often not required to know the full graph. Hence, by marginalising the posterior we can uncover the confidence we have about a specific causal effect, which may already fall below a specified tolerance level. Having such a model is very desirable as interventions are hard to perform, sometimes unethical and even impossible \citep{peters2017elements}. 

A key component in learning causal structures is the identifiability under the availability of observational data. While some assumptions are always required to say anything about the underlying causal process, additional assumptions are sometimes made to especially make the causal model identifiable from (observational) data. These additional assumptions are not necessarily part of the data generating process, and hence the recovered causal structure may be incorrect due to model misspecification. In addition, identifiability results are usually asymptotic in the number of samples. Causal discovery in a limited sample regime calls for active learning to perform interventions to improve identifiability. Such setups benefit from probabilistic reasoning about unknown causal structures. We are interested in such settings and we propose a Bayesian framework for linear SCM's with additive noise which can quantify the uncertainty in the learned causal structure. 

Our contributions are as follows:
\begin{itemize}[leftmargin=*]
    \item We perform Bayesian inference over the unknown causal structures. The posterior over this distribution is intractable. Therefore, we perform variational inference and demonstrate how we can model the variational family over this distribution. 
    \item The key contribution is to model distributions over adjacency matrices of the causal structures with an autoregressive distribution using an LSTM. 
    \item Empirically demonstrate the performance of our modeling choice. 
    
    \item Evaluating Bayesian causal inference techniques are hard in practice. We discuss the difficulty entailed in evaluation, as well as provide insights which alleviate this problem. 
\end{itemize}

\section{Problem Setting}

\paragraph{Causal modeling}
Consider a set of $d$ random variables $\mathbf{X} := \{X_1,\dots,X_d\}$. A Structural Causal Model (SCM) \citep{peters2017elements} over $\mathbf{X}$ is defined as set of structural assignments
\begin{equation}\label{eq:scm}
X_i := f_i(\X_{\pi_\G(i)},\epsilon_i)  ~, i=1, ..., d
\end{equation}
corresponding to a Directed Acyclic Graph (DAG) $\G$ with vertices $\mathbf{X}$. In here,  $\pi_\G(i)$ are the parents of $X_i$ in $\G$, $\epsilon_i$ are independent noise variables with probability density $P_{\epsilon_i}$ and the $f_i$'s are (potentially non-linear) functions. The SCM entails a joint probability distribution $P_\X$ over the random variables. In this work, we assume that all the endogenous variables are observed, that is, there are no hidden confounders.

A popular instantiation of this generic framework are linear SCMs with additive noise, given by:
\begin{equation}\label{eq:additive_scm}
x_i := \btheta_i^T \X_{\pi_\G(i)} + \epsilon_i   ~~~~ \btheta_i \in \R^{|\pi_\G(i)|} 
\end{equation}
wherein $\btheta_i$ are parameters (edge weights) of the linear functions $f_i$. Alternatively, this can be written as
\begin{equation}\label{eq:additive_scm_ad}
x_i := \btheta_i^T \left(\mathbf{X}\circ \A_{\G_i}\right)+ \epsilon_i   ~~~~ \btheta_i ~,~ \A_{\G_i}\in \R^{d} ~,
\end{equation}
where $\circ$ corresponds to elementwise product, $\A_{\G_i}$ is the $i$\textsuperscript{th} row of $\A_\G\in \R^{d\times d}$, the $(0,1)$-adjacency matrix of $\G$. We restrict our further exposition to continuous random variables. However, the framework presented in the remainder of the paper applies to discrete random variables as well.
We focus on linear SCMs with Gaussian variables, as they are non-identifiable in this setting and hence obtaining uncertainty estimates is useful. Non-linear additive noise SCM's with Gaussian variables are identifiable, and hence a single point estimate suffices.


%

\paragraph{NOTEARS - Causal inference as continuous optimization problem} In practice, we often have data while the causal structure that underlies the data generating process is unknown to us. Causal discovery is concerned with recovering this causal structure, e.g. in the form of an SCM, from observational data or interventional data (or both). Assuming a linear SCM, this coincides with estimating the graph $\G$, i.e. its adjacency matrix $\A$, and its corresponding edge weights $\btheta=\{\btheta_1, ..., \btheta_d\}$, given data $\gD := \{\bx_1, ..., \bx_n\} \in \gX^n$.

In score-based causal discovery, we use a score function $F: \{0,1\}^{d \times d} \times \gX^n \mapsto \R$ to find the graph that best corresponds to the data as follows:
\begin{equation} \label{eq:score_based_dag_learning}
\argmin_{\A \in \{0,1\}^{d \times d}} F(\A, \gD) \quad \text{s.t.} ~~ \gG(\A) \in \operatorname{DAG}(d)
\end{equation}
where $\A$ is the $\{0,1\}$ adjacency matrix of graph $\gG$.

However, since the search space of DAGs of this combinatorial optimization problem grows in the order of $\mathcal{O}(2^{d^2})$, solving (\ref{eq:score_based_dag_learning}) becomes infeasible even for a relatively small number of variables $d$. 

Addressing this issue, \citet{zheng2018dags} propose an alternative formulation that converts the combinatorial problem into the following continuous program:
\begin{equation} \label{eq:notears}
\argmin_{\mW \in \R^{d \times d}} F(\mW, \gD) \quad \text{s.t.} ~~ \text{tr} \left( e^{\mW} \right) - d = 0 ~.
\end{equation}
that can be solved with standard tools of constrained optimization such as the Lagrange method.
In here, we have converted the adjacency matrix $\A \in \{0, 1 \}^{d \times d}$ into a weighted adjacency matrix $\mW \in \R^{d \times d}$ such that $a_{i,j} = 1 \Leftrightarrow w_{i,j} \neq 0$. 


\paragraph{Bayesian inference over DAGs} An important aspect towards acting intelligently upon the knowledge about the causal structure which has been inferred from finite data demands reasoning about its uncertainty. For instance, in scientific inquiry, planning \emph{interventional experiments} to find out more about the causal mechanisms that govern a data-generating system of interest requires quantifying \emph{epistemic uncertainty} over DAGs. Bayesian inference gives us a principled framework for the treatment of such uncertainty (Figure~\ref{fig:vi}).

In the Bayesian framework, we encode our prior structural knowledge in form of a \emph{prior} $p(\gG)$ over DAGs. By combining the prior with the \emph{likelihood} $p(\gD| \gG)$ through Bayes' rule, we obtain a \emph{posterior distribution} over DAGs:
\begin{equation}\label{eq:true_posterior}
    p( \G | \gD) = \frac{p(\gD|\G) p(\G)}{p(\gD)}
\end{equation}
where $p(\gD) = \sum_{\G} \int_\theta p(\X|\G, \btheta) p(\btheta | \G) p(\G) d\btheta$ is the \emph{model evidence}.
A key issue with (\ref{eq:true_posterior}) is the intractable evidence term in the denominator. Even if the integral over $\btheta$ can be solved analytically such as in the case of linear structural equations with Gaussian prior and likelihood with standard parameter priors~\citep{geiger2002parameter}, the sum over possible DAGs grows in the order of $\gO(2^{d^2})$, making its computation infeasible even for a small number of variables $d$.


%


\section{Variational Causal Networks}
\begin{figure}
    \centering
    \includegraphics[height = 5.5cm]{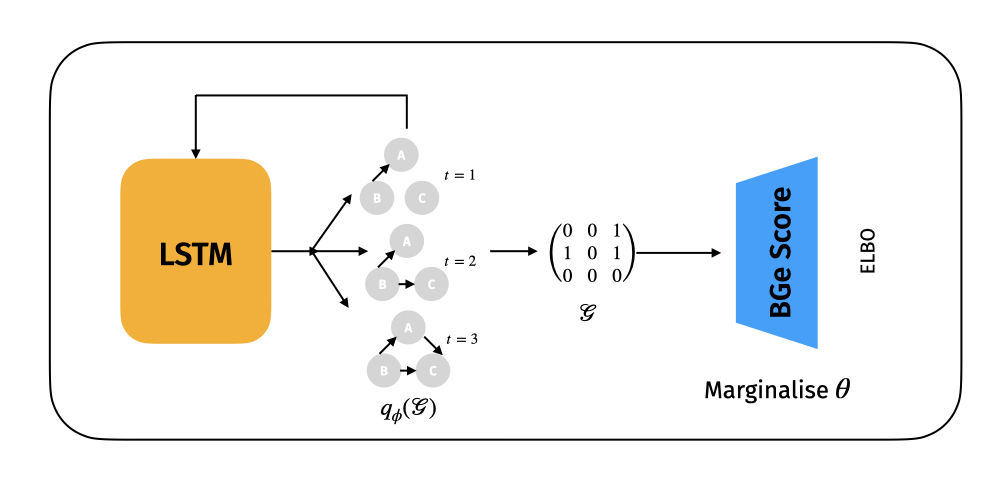}
    \caption{Schematic Diagram of Variational Causal Networks}
    \label{fig:main}
\end{figure}

In this section, we present our variational inference (VI) framework for approximating Bayesian posteriors over causal structures (Equation~\ref{eq:true_posterior}). First, we derive the evidence lower bound and sketch out how to perform VI on DAGs. Unlike the variational inference in a latent variable model~\citep{blei2017}, the evidence lower bound is still intractable in the case of causal models due to the superexponential number of graph configurations. Therefore, we introduce a novel parametric variational family which makes the intractable ELBO into a tractable one while still having the flexibility to model complex distributions over DAGs. Such a variational family can be modelled using autoregressive models like LSTM~\citep{hochreiter1997long}. This particular instantiation of approximate Bayesian inference for SCMs with the presented variational family is called Variational Causal Networks (VCN).

\subsection{Variational Inference for Causal Structures}
In variational inference \citep{blei2017}, we aim to approximate the Bayesian posterior $p(\G | \gD)$ by a variational distribution $q_\phi(\gG)$, parameterized by $\phi$, that has a tractable density. 
To learn the parameters of our variational distribution, we minimize the KL-Divergence between $q_{\phi}(\G)$ and the true posterior $p(\G|\gD)$ which is equivalent to maximising the \emph{Evidence Lower Bound (ELBO)}, given by the following proposition.
\begin{proposition}\label{prop:elbo_scm} \textbf{(ELBO)}
Let $q_{\phi}(\G)$ be the variational posterior over causal structures. Then the evidence lower bound (ELBO) is given by:
\begin{equation}\label{eq:elbo}
\begin{split}
    \log p(\gD) \geq  \mathcal{L}(\phi;\gD) = \E_{q_\phi(\G)}\left[\log p(\gD|\G)\right] - \kl\left(q_\phi(\G)||p(\G)\right) 
\end{split}
\end{equation}
\end{proposition}

To maximise the ELBO, we estimate the gradients $\nabla_{(\phi)}\mathcal{L}(\phi;\gD)$ to employ a form of gradient ascent on the objective. 

\subsection{A Variational Family for Causal Structures}
A key challenge to variational inference over DAGs concerns the choice of the variational distribution $q_{\phi}(\G)$ over discrete DAGs $\gG$. For instance, if we choose $q_\phi(\gG)$ naively as categorical distribution over the space of DAGs, the dimensionality of $\phi$ grows super-exponentially in the number of variables $d$ (i.e. $\mathcal{O}(2^{d^2})$), rendering this choice impractical. Hence, we need to find a way of representing $q_\phi(\gG)$ that is not combinatorial in its nature while ensuring a rich family of distributions over graphs. In addition, to be able to compute Monte Carlo gradient estimates of the ELBO in (\ref{eq:elbo}), $q_\phi(\G)$ needs to have a tractable probability density.

Aiming to fulfil these requirements, we represent graphs by their adjacency matrix $\A \in \{0,1\}^{d \times d}$ and model $q_\phi$ as a discrete distribution over such $\{0,1\}$-matrices. While \citet{ke2019learning} represent each entry of the adjacency matrix as independent Bernoulli variable, such approach is insufficient to capture any dependencies between the entries of the adjacency matrix which are necessary to ensure that $q(\A)$ only assigns positive probabilities to adjacency matrices corresponding to DAGs. For instance, to avoid cycles of length two, $(\A)_{i,j}$ needs to have a strong negative dependency $(\A)_{j,i}$, i.e., if there is a directed edge $i \rightarrow j$ there must be no edge $j \rightarrow i$, and reverse. In addition, a factorisable distribution can only capture unimodal distributions while the posterior over causal structures in the non-identifiable case could have exponential number of modes. Hence, the variational distribution needs to be parameterised such that multiple modes could be captured. In order to address these issues, we model this discrete distribution as an autoregressive distribution over entries of the adjacency matrix using an LSTM~\citep{hochreiter1997long}. 

Let $q_\phi(\G)=q_\phi(\A_{\G})$ be defined in an autoregressive manner as follows:
\begin{equation*}
    \begin{split}
         q_\phi(\A_{\G}) &= \prod_{i=1}^{d(d-1)} q_{\phi}(a_{\G_i}|a_{\G_{1:i-1}})\\ 
         s.t. ~~~~ q_{\phi}(a_{\G_i}|a_{\G_{1:i-1}}) &:= \text{Bernoulli}\left(a_{\G_i}; f_\phi\left(a_{\G_{1:i-1}}\right)\right),~~\A_{\G} = \{a_{\G_i}\}_{i=1}^{d(d-1)}
    \end{split}
\end{equation*}
where only the non-diagonal elements of the adjacency matrix are modelled and $f_\phi$ is a function which predicts the Bernoulli parameter based on the previous realisations of the entries of the adjacency matrix, thus making the distribution autoregressive. We model this function using an LSTM~\citep{hochreiter1997long}. 

Note that using an autoregressive distribution on the entries of the adjacency matrix helps to implicitly keep a distribution over super-exponential number of graphs, and also be able to sample from that distribution. 

\paragraph{Prior over graph structures} Choosing the appropriate prior $p(\G)$ over graph structures is important to learn good approximation of the posterior. In causal discovery, it is usually a requirement that the search space of all graphs is a DAG. However, in the parameterisation described above, the discrete distribution is defined over all possible graphs, including the one which contains cycles. Having support over graphs with cycles and using maximum likelihood estimation results in high probability mass corresponding to a fully connected graph in the approximate posterior. Therefore, appropriate DAG regularizers are required. We employ the result of NOTEARS~\citep{zheng2018dags} to define a Gibbs distribution as the prior which helps us to limit the support of graphs to just DAGs. We can also encode additional assumptions about the data generating process such as sparsity. The Gibbs distribution in this case could be defined as:
\begin{equation}
    p(\G) \propto \exp(-\lambda_t g(\A_{\G})- \lambda_{s} \norm{\A_{\G}}_1)
\end{equation}
where $g(\A_{\G})$ is the DAG constraint given by the matrix exponential~\citep{zheng2018dags}, i.e $g(\A_{\G}) = \tr\left[e^{\A_{\G}}\right] - d$. Matrix binomial can also be used to define $g$, as given in \cite{yu2019dag}. The second term in the Gibbs distribution corresponds to the sparsity constraint. $\lambda_t$ and $\lambda_s$ are tunable hyperparameters. Typically, $\lambda_t \to \infty$ if graphs with cycles should have zero probability mass under the posterior. However, any large value should still give a reasonable prior with very low probabilities for cyclic graphs. 

\paragraph{Marginal Likelihood} We use standard parameter priors which ensure marginal likelihood $\log p(\gD|\G)$ is "score-equivalent", i.e. all the DAGs which are in the same MEC have the same marginal likelihood. This can be ensured with a Gaussian-Wishart prior over the parameters $\btheta$ and the marginalisation can be done in closed form~\citep{geiger2002parameter,kuipers2014addendum} (called as the \emph{BGe score}). Using this parameter prior to calculate marginal likelihood in the ELBO makes the ELBO score-equivalent. Details are given in Appendix~\ref{app:bge}.

\subsection{Estimating Gradients}

Since the KL regulariser is defined on a discrete distribution with many parameters, obtaining a Monte Carlo estimate of this term (and hence the corresponding ELBO) requires unbiased gradients of $\phi$. Therefore, we use the score-function gradient estimator~\citep{williams1992simple} with exponential moving average baseline for obtaining the gradients. As the score function estimator does not require the estimand function to be differentiable, we can perform closed form marginalisation of the parameters with the standard parameter priors like that of Gaussian-Wishart.

The detailed algorithm is given in Algorithm~\ref{algo:vcn}.

\begin{algorithm}
\caption{VCN Algorithm}
\label{algo:vcn}
\begin{algorithmic}[1]
 \State $\phi \gets$ initialise parameters
 \Repeat
 \State Sample $L$ graphs $\{\G^{(i)}\}_{i=1}^{L}$ autoregressively from $q_\phi(\A_{\G})$ using an LSTM.
 \State Calculate the per-sample marginal log-likelihood $\log p(\gD|\G^{(i)})$ by marginalising $\btheta$ to obtain the BGe Score~\citep{geiger2002parameter,kuipers2014addendum}.
 \State $g \gets \frac{1}{L}\sum_{i=1}^{L} \nabla_{\phi} \left[\log p(\gD|\G^{(i)}) - \left[\log q_\phi(\G^{(i)})- \log p(\G^{(i)})\right]\right]$
 \State $\phi\gets$ Update parameters using gradients $g$. 
\Until $\phi\gets$ converge.
\end{algorithmic}
\end{algorithm}
\section{Related Work}
\textbf{Causal Discovery:} Broadly, causal structure learning approaches can be mainly categorised to two different paradigms with regards to kind of data used: methods involving learning structure from purely observational data and methods involving learning structure from both observational and interventional data. Both these approaches strongly rely on the identifiability of SCM.~\citet{zheng2018dags,yu2019dag,lachapelle2019gradient} use a score-based objective with global search over DAG space to learn the structure.~\citet{lachapelle2019gradient} extend the linear model of~\cite{zheng2018dags} to non-linear case using neural networks. GES~\citep{chickering2002optimal,ramsey2017million} greedily searches over the space of CPDAG's and maximize a score based on Bayesian Information Criterion (BIC). LiNGAM~\citep{shimizu2014lingam} uses an Independent Component Analysis (ICA) based approach by making the assumption that either the variables or one of the noise variables are non-Gaussian. PC~\citep{spirtes2000causation} algorithm uses a constraint based approach for causal discovery.
 
 Different from the above approaches, a few approaches use a combination of observational and interventional data to learn the causal structure~\citep{JMLR:v13:hauser12a}. ICP~\citep{peters2016causal} assumes a linear SCM and resorts to conditional independence testing to test the hypothesis of invariance, a concept wherein the plausible causal predictors of a given variable are stable across different interventions.~\citep{heinze2018invariant} extend ICP to non-linear SCMs. These methods do not infer the complete graph but instead just the causal parents of a particular variable. The difficulty of these approaches rest in the fact that conditional independence tests are hard to perform. More recently,~\citep{bengio2019meta, ke2020amortized} and~\citep{ke2019learning} use a meta-learning objective to learn the causal structure using interventional data. 

\textbf{Bayesian Approach to Structure Learning:} Existing approaches for Bayesian learning of causal graphs are mainly concerned with efficient sampling of graph structures from the posterior distribution. A popular choice for such an approach is the Markov Chain Monte Carlo (MCMC) with suitably chosen energy functions and heuristics~\citep{ellis2008learning,niinimaki2016structure,madigan1995bayesian,kuipers2017partition,heckerman1999bayesian, friedman2003being}. While \citet{niinimaki2016structure} is concerned with causal discovery by sampling graph structures from partial orders of edge orientation, \citet{madigan1995bayesian} proposes an approach for learning probabilistic graphical models with discrete variables called as structure MCMC. \citet{kuipers2017partition,grzegorczyk2008improving} and \citet{ellis2008learning} propose improved MCMC techniques for sampling graph structures by restricting graph space to certain topology and sampling from a restricted space. \citet{heckerman1999bayesian} use a BIC based scoring function to learn a Bayesian network by sampling from Metropolis-Hastings under different energy configurations. An estimate of MAP is obtained using Maximum Likelihood to get the final graph structure. \citet{agrawal2019abcd} use DAG bootstrapping to estimate the posterior over graphs and use this for budgeted experimental design for causal discovery. While all these techniques are intended for a Bayesian approach to structure learning, they face the problem of efficiently approximating the posterior over graph structures and hence resort to either heuristics or restricted graph structures. However, we do not make any simplifying assumptions about the graph structures and we can efficiently approximate the posterior while being able to sample exactly from them.

\section{Experiments}
To validate the modelling choice presented in the previous section, we perform experiments mainly focusing on the following aspects: (1) since we approximate the posterior, we evaluate how close is the approximation to the true posterior in lower dimensional settings ($\leq 4$) where enumeration of true posterior is possible. (2) We outline the difficulty involved in evaluating the Bayesian Causal inference models in higher dimensions where enumeration is not possible and suggest possible metrics which alleviate the problem and quantify how well the true posterior is approximated. We further evaluate our model on these metrics.

\paragraph{Key Findings}
We summarise the experimental results as follows: (1) The proposed autoregressive approach of VCN performs better than a factorised distribution~\citep{ke2019learning} in all the considered settings. This is due to the ability of an autoregressive distribution to model multi-modal posteriors as compared to the unimodal factorised distribution. (2) In higher dimensions, the proposed approach compares favourably to competitive baselines on various metrics. (3) VCN achieves a good approximation vs run-time trade-off, thus making it a suitable approach for Bayesian causal inference.  

\paragraph{Experimental Settings}\label{sec:exp_setting}
We evaluate our method on both synthetic and real datasets. For generating synthetic data, we follow the procedure of NOTEARS~\citep{zheng2018dags}. We sample a DAG at random from an Erdos-Renyi (ER) model with expected number of edges equal to $d$. 
Each reported result is over 20 different random graphs. The models were trained with a learning rate of $1e-2$ using the Adam optimiser~\citep{kingma2014adam} for 30k epochs. We consider the settings when we have $n=10, 100$ samples. For taking the Monte Carlo estimate of the ELBO, we take $L=1000$ samples. $\lambda_s$ is fixed to 0.01 and $\lambda_t$ is annealed from 10 to 1000 with an exponential annealing schedule. 

\paragraph{Baselines} Parameterising the posterior with a \textbf{factorised distribution}~\citep{ke2019learning} is our main baseline as it involves differentiable causal learning based techniques similar to our approach. We also compare with DAG Bootstrap~\citep{agrawal2019abcd} where the posterior is estimated by bootstrapping the data with any causal discovery algorithm and then forming an empirical estimate of the posterior based on frequency count. We use LiNGAM~\citep{shimizu2014lingam} and NOTEARS~\citep{zheng2018dags} as underlying causal discovery algorithms for DAG Bootstrap. They are denoted by \textbf{Boot Lingam} and \textbf{Boot Notears} respectively. In addition, we compare with the MCMC approach of Minimal \textbf{IMAP MCMC}~\citep{agrawal2018minimal}. Details of experimental settings of each of these techniques is given in Appendix~\ref{app:exp_base}.

\subsection{Evaluation Metrics}
For low dimensional variables $(d\leq 4)$, we can enumerate the true posterior for all graphs. Therefore, in this setting we compute the distance between the variational posterior learned from data and the true posterior.

For higher dimensional variables, where enumerating the true posterior is not possible, evaluating Bayesian causal inference algorithms is not straightforward. Directly comparing the likelihoods do not necessarily ensure that the posterior learned is good, as graphs with more edges always have higher likelihoods. Instead, we employ the following metrics: 
\paragraph{Expected Structural Hamming Distance ($\E[\text{SHD}]$)} Given that the true posterior has modes over all the graphs inside the MEC corresponding to the ground truth graph, we can sample the graphs from the model and then compute the Structural Hamming Distance (SHD) between these samples and the ground truth. We can then compute the empirical mean of these SHDs as a metric. As all the graphs inside the MEC have the same number of edges, this metric indicates how well the approximated posterior is close to the ground truth on average. If $\G_{\text{GT}}$ is the ground truth data generating graph, then $\E[\text{SHD}]$ is given by
    \begin{equation}
         \E_{q_\phi(\G)}[\text{SHD}] \approx \frac{1}{T}\sum_{i=1}^T\left[ \text{SHD}(\G^{(i)},\G_{\text{GT}})\right]~~~~\G^{(i)}\sim  q_\phi(\G)
    \end{equation}
    
\paragraph{Area Under Reciever Operating Curve (AUROC)} If we care for just feature probabilities of quantities like the presence of an edge, we can compute edge beliefs and compute the Receiver Operating Curve (ROC). The area under this curve can be treated as a metric for evaluating the Bayesian model. However, it does not necessarily give a full picture of the multiple modes of the posterior and whether the sampled graph is far away /close to the MEC of the true graph. Nevertheless, it can still be informative of the edge beliefs of the learned approximation. Details of computing this is give in~\citep{prill2010towards}.

We would like to note that though these metrics capture to a reasonable extent how well the true posterior is approximated, they are still imperfect. Since the true posterior has mass over all the graphs corresponding to MEC of the ground truth graph, the true posterior does not necessarily evaluate to zero on $\E[\text{SHD}]$ and one on AUROC. Hence, evaluating the learned model on these metrics only partially indicates how well the true posterior is approximated. 

\subsection{Results}
\begin{figure}[t!]
    \begin{minipage}{0.5\textwidth}
    \centering
       \includegraphics[height=4.0cm]{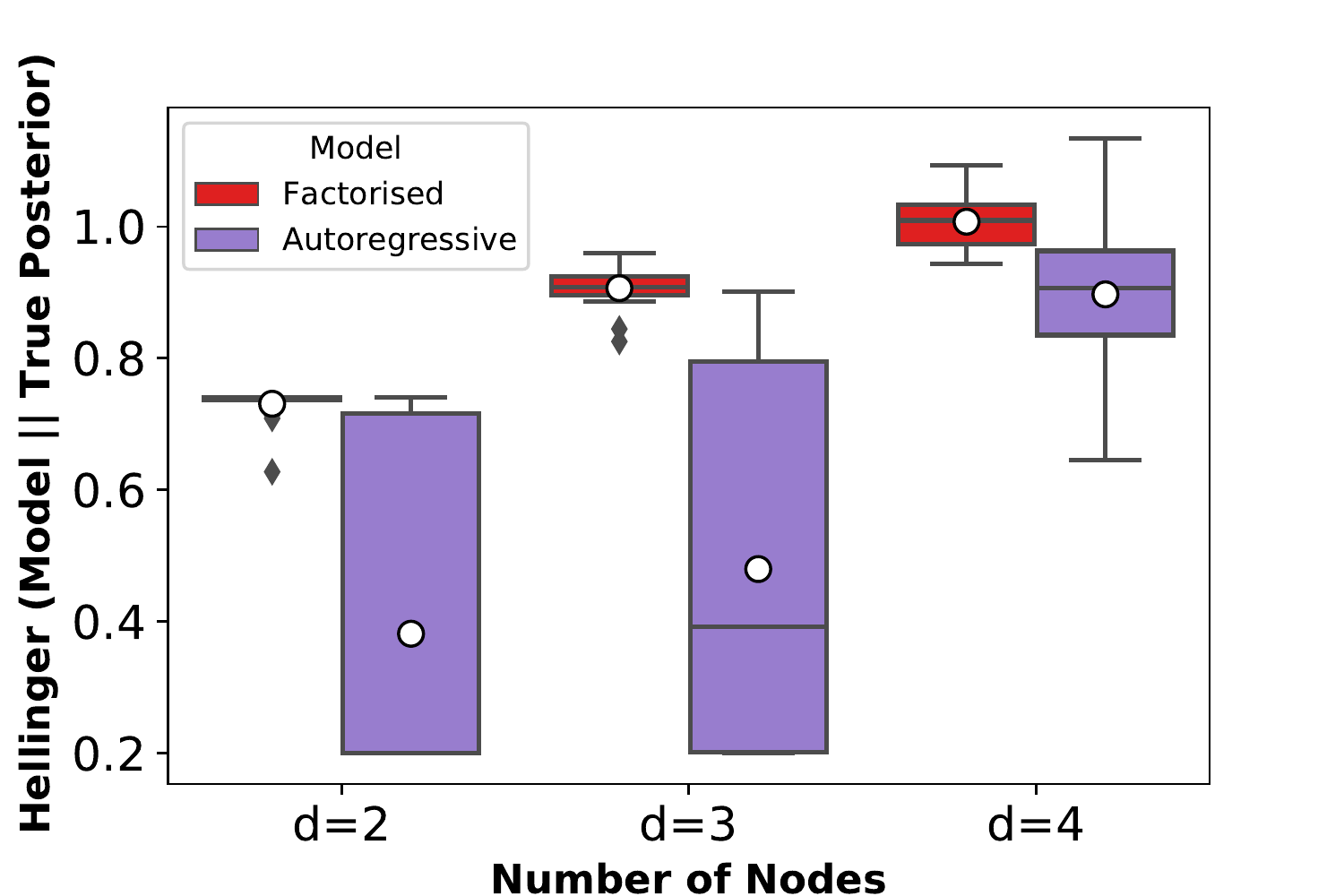}
    \caption{Hellinger distance of the full posterior of the approximation with the true posterior. }
    \label{fig:kl_tp} 
    \end{minipage}%
    ~
    \begin{minipage}{0.5\textwidth}
    \centering
    \includegraphics[height=4.0cm]{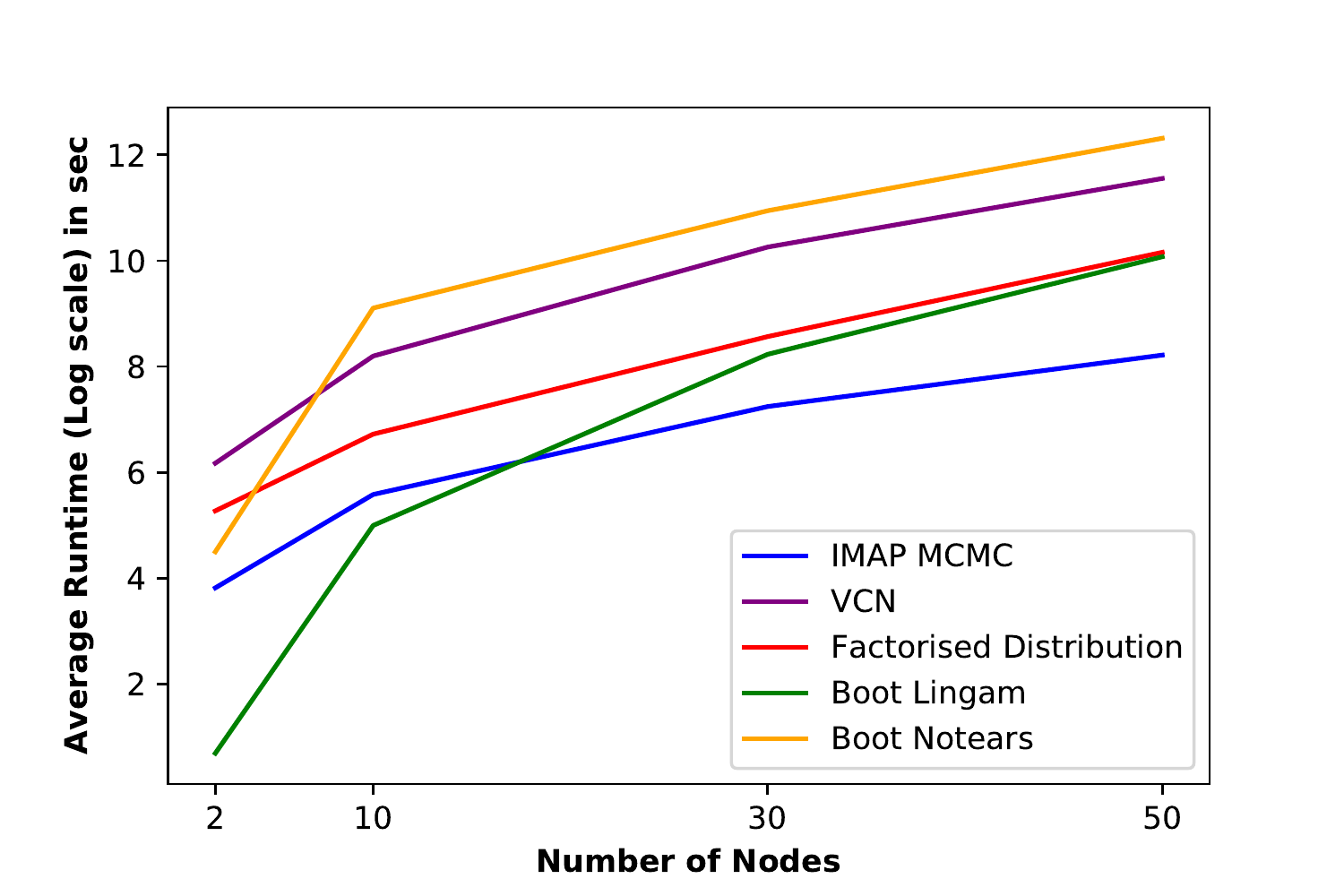}
     \caption{Average runtime (in seconds, log scale) of different Bayesian causal inference approaches.}
     \label{fig:runtime}
    \end{minipage}%
\end{figure}
\subsubsection{Estimation of True Posterior}
As indicated before, when the number of variables in the graph is small, we can enumerate the true posterior and compute the divergence/distance between the approximation and the true posterior. Figure~\ref{fig:kl_tp} presents the Hellinger distance of the approximation up to four nodes. It can be seen that the autoregressive distribution of VCN reconstructs the true posterior much better than the factorised distribution, mainly due to the non-identifiability of graphs and the fact that the true posterior is multimodal.
\subsubsection{Evaluation in higher dimensions}
\begin{figure}[t!]
    \centering
    \begin{subfigure}[t]{0.5\textwidth}
        \centering
        \includegraphics[height=1.9in]{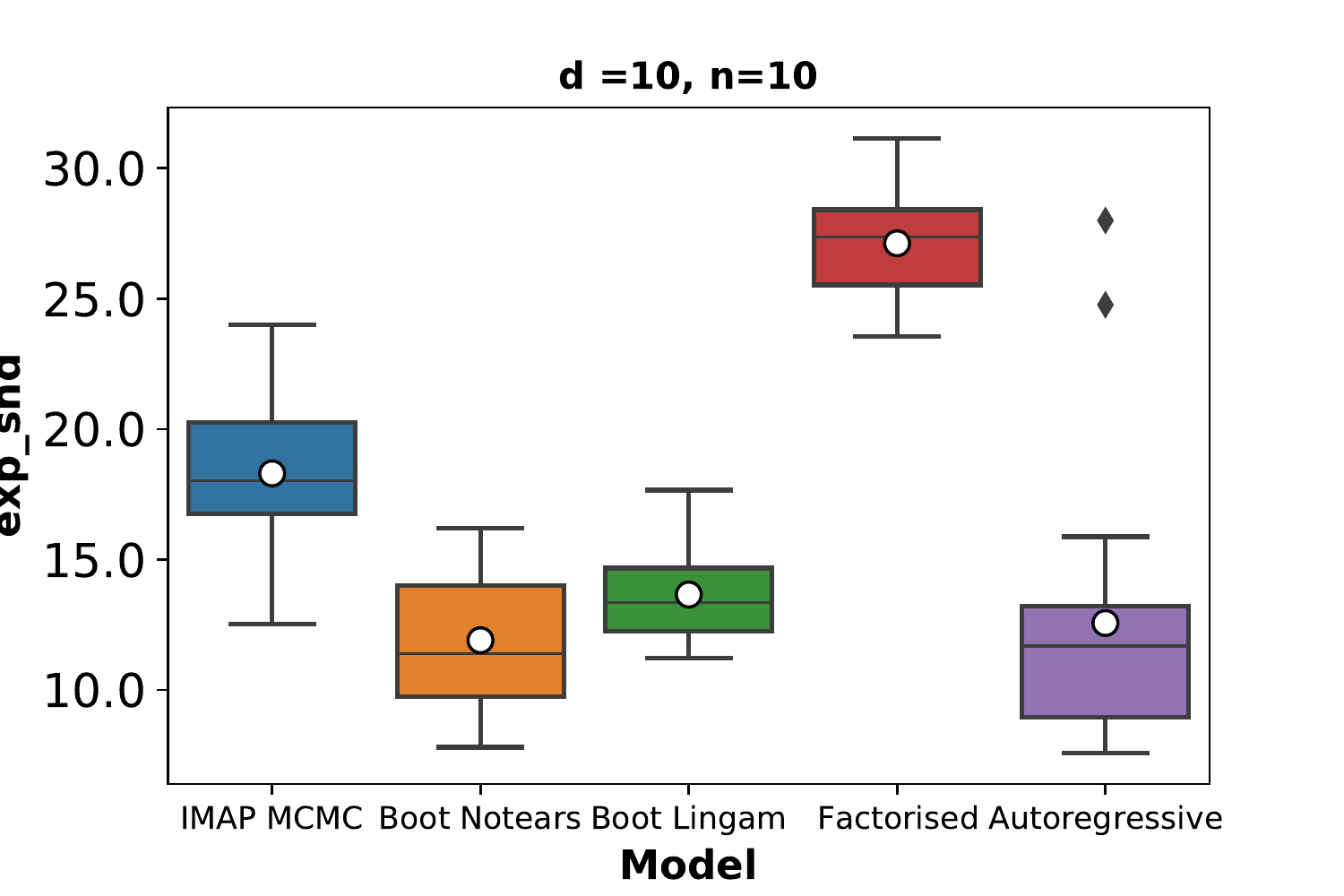}
    \end{subfigure}%
    ~ 
    \begin{subfigure}[t]{0.5\textwidth}
        \centering
        \includegraphics[height=1.9in]{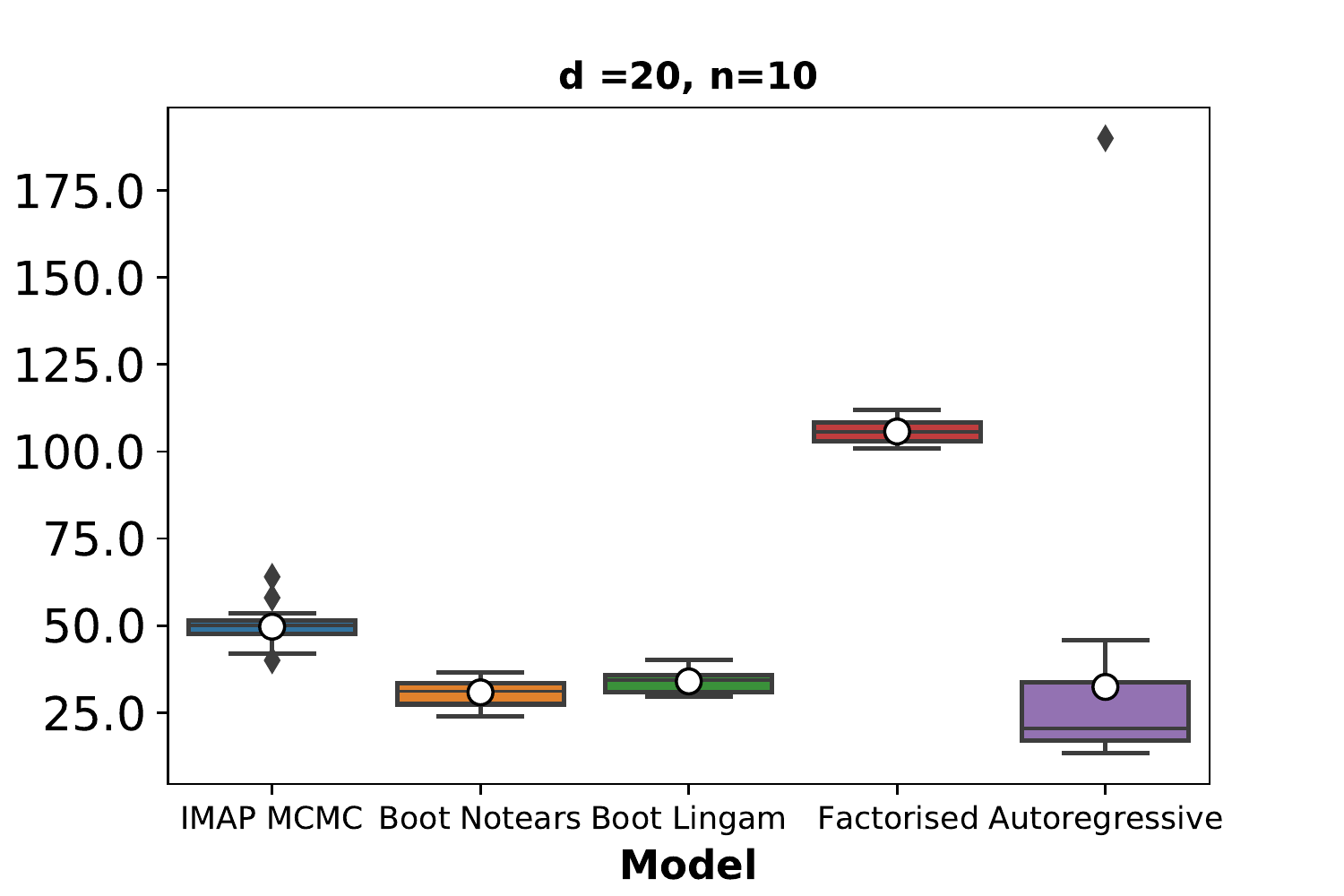}
    \end{subfigure}\\
    \begin{subfigure}[t]{0.5\textwidth}
        \centering
        \includegraphics[height=1.9in]{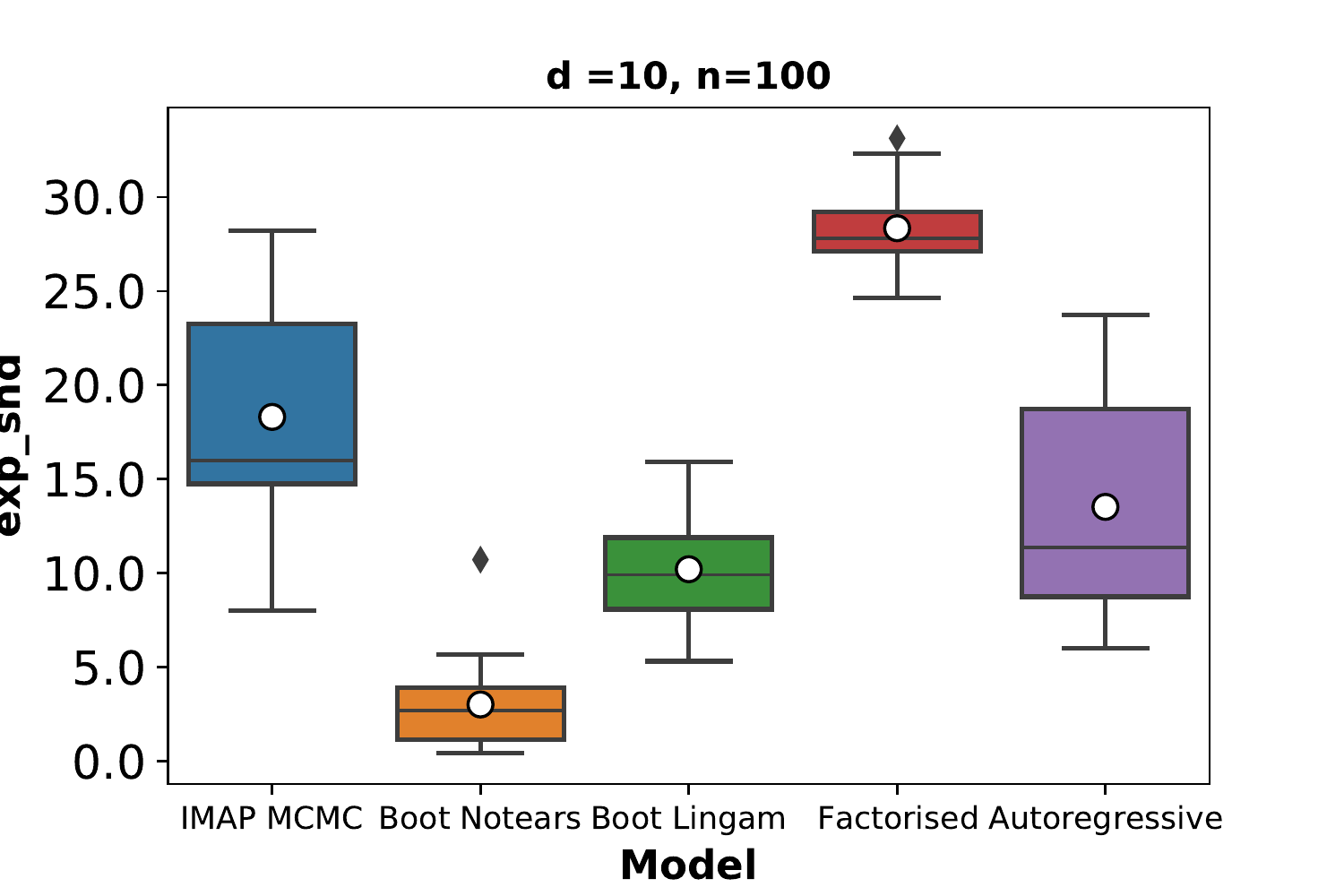}
    \end{subfigure}%
    ~ 
    \begin{subfigure}[t]{0.5\textwidth}
        \centering
        \includegraphics[height=1.9in]{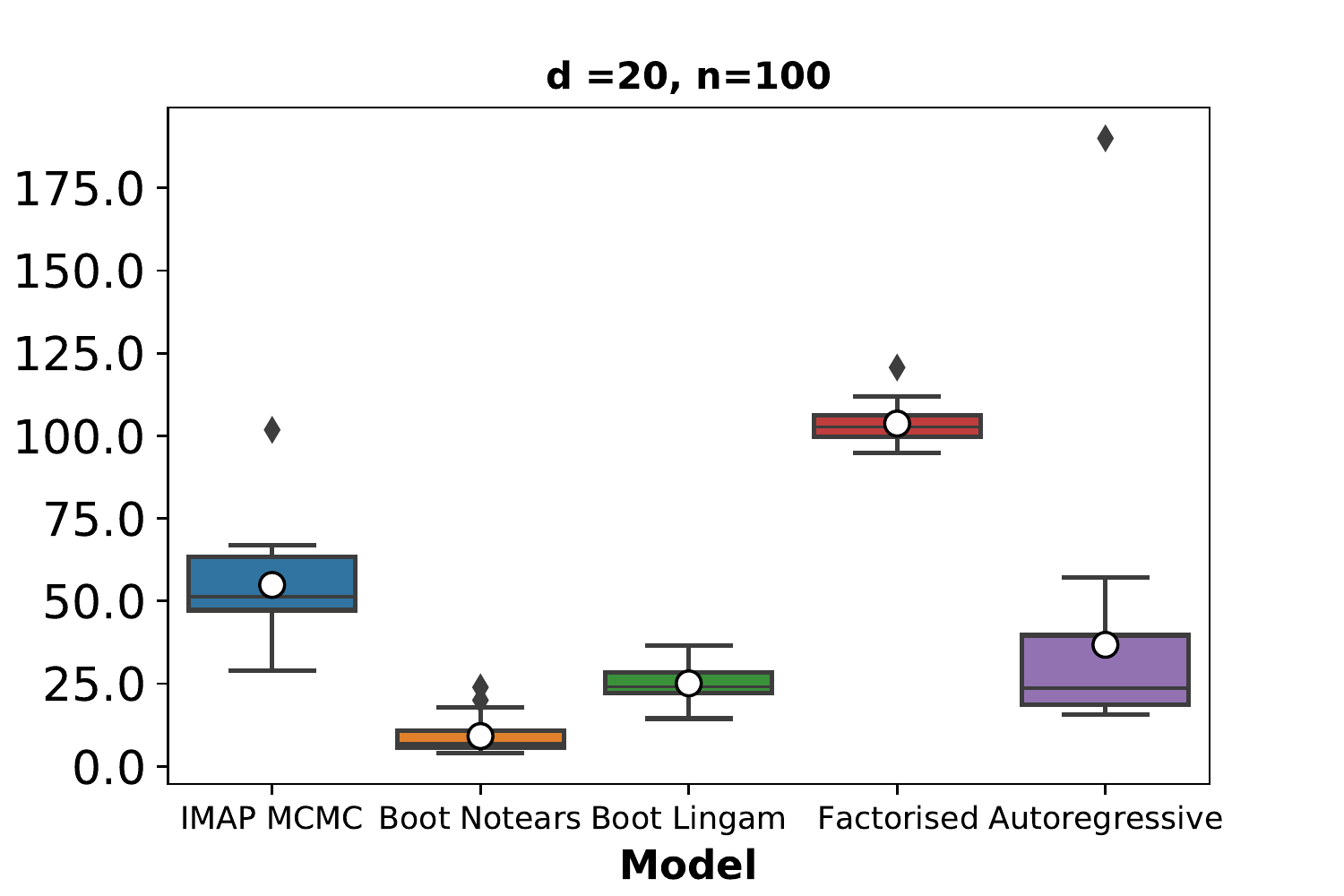}
    \end{subfigure}
    \caption{$\E\left[\text{SHD}\right]$ for $d=10$ and $d=20$ node ER random graphs (lower is better). Results obtained using 20 different random graphs.}
    \label{fig:exp_shd}
\end{figure}

\begin{figure}[t!]
    \centering
    \begin{subfigure}[t]{0.5\textwidth}
        \centering
        \includegraphics[height=1.9in]{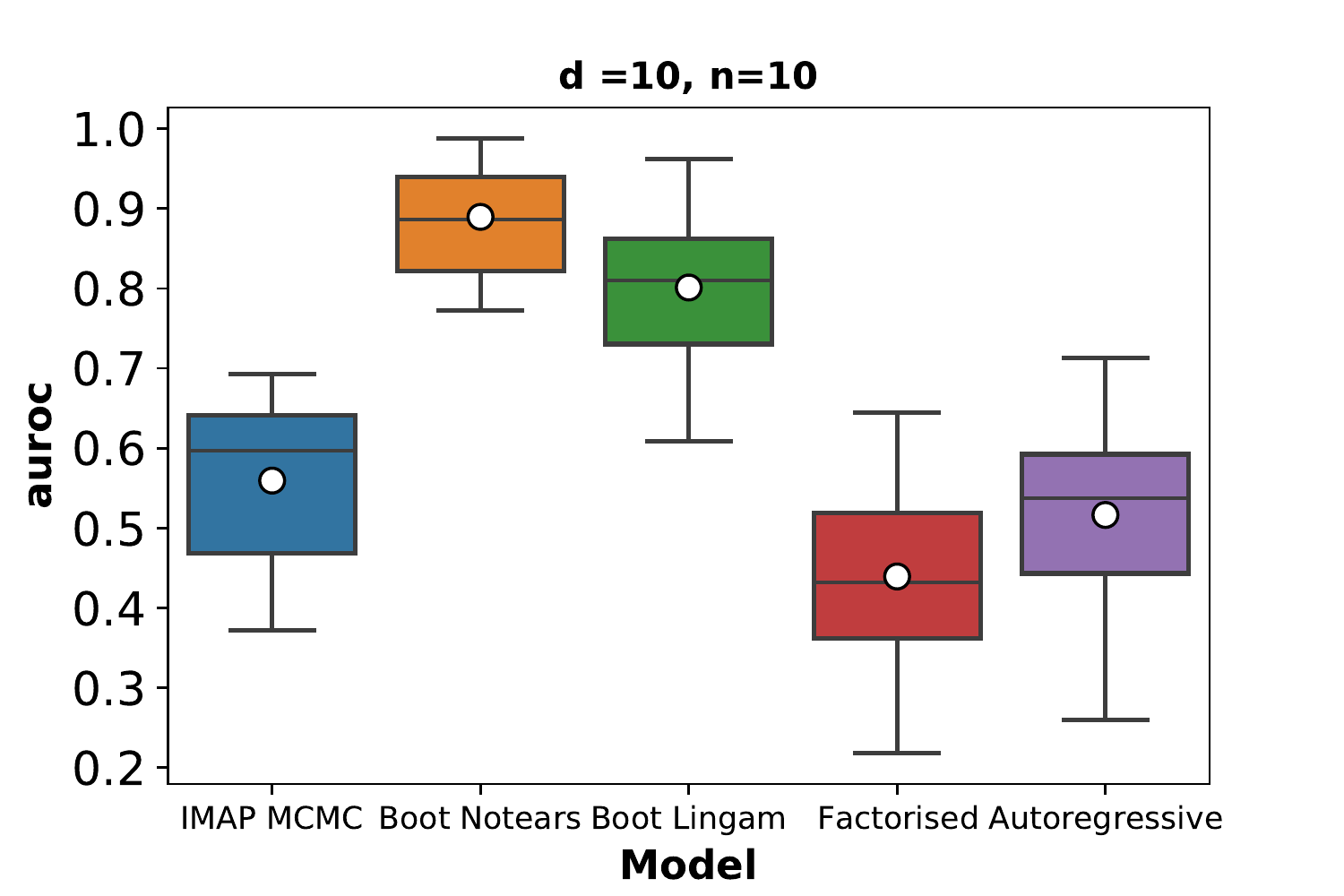}
    \end{subfigure}%
    ~ 
    \begin{subfigure}[t]{0.5\textwidth}
        \centering
        \includegraphics[height=1.9in]{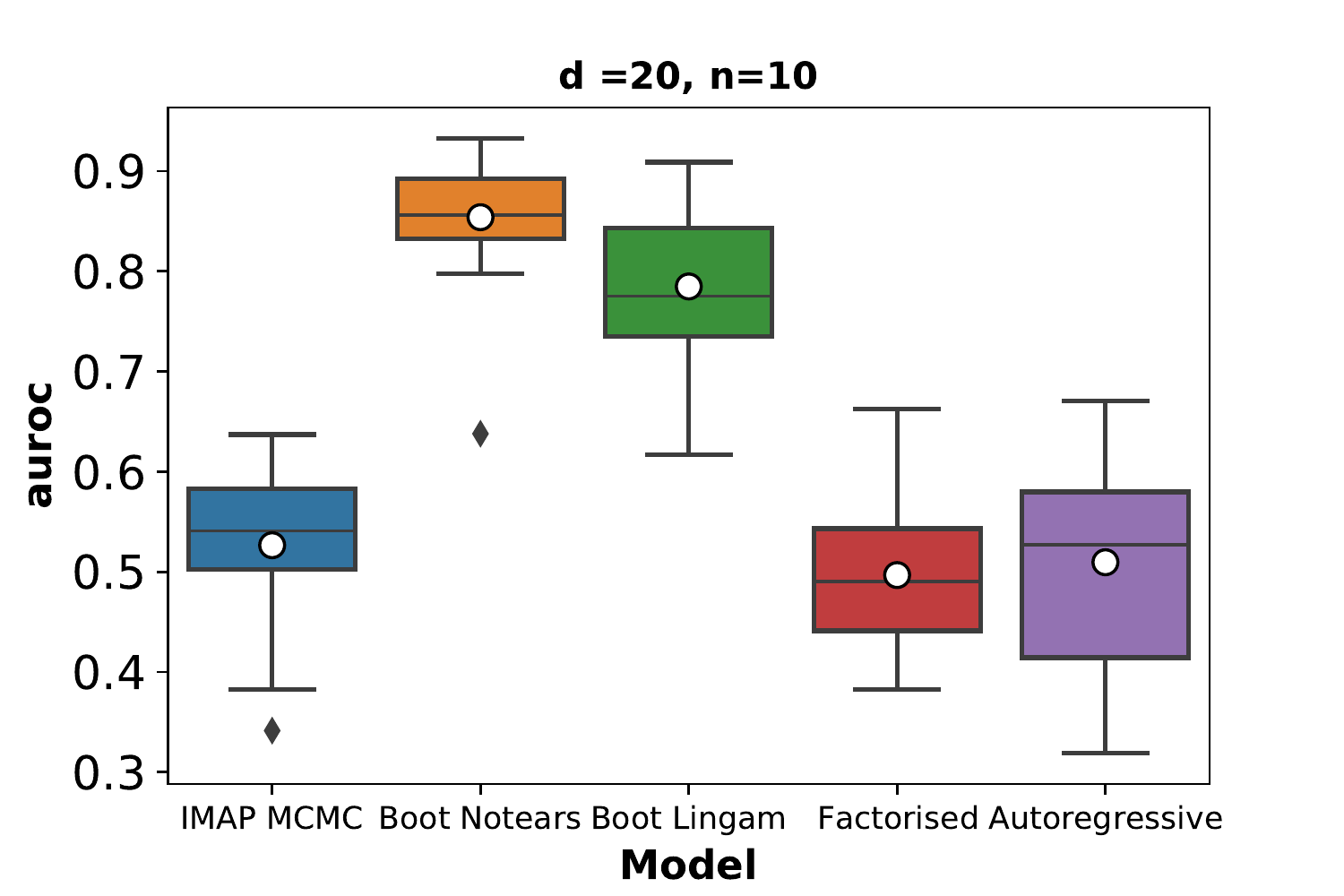}
    \end{subfigure}\\
    \begin{subfigure}[t]{0.5\textwidth}
        \centering
        \includegraphics[height=1.9in]{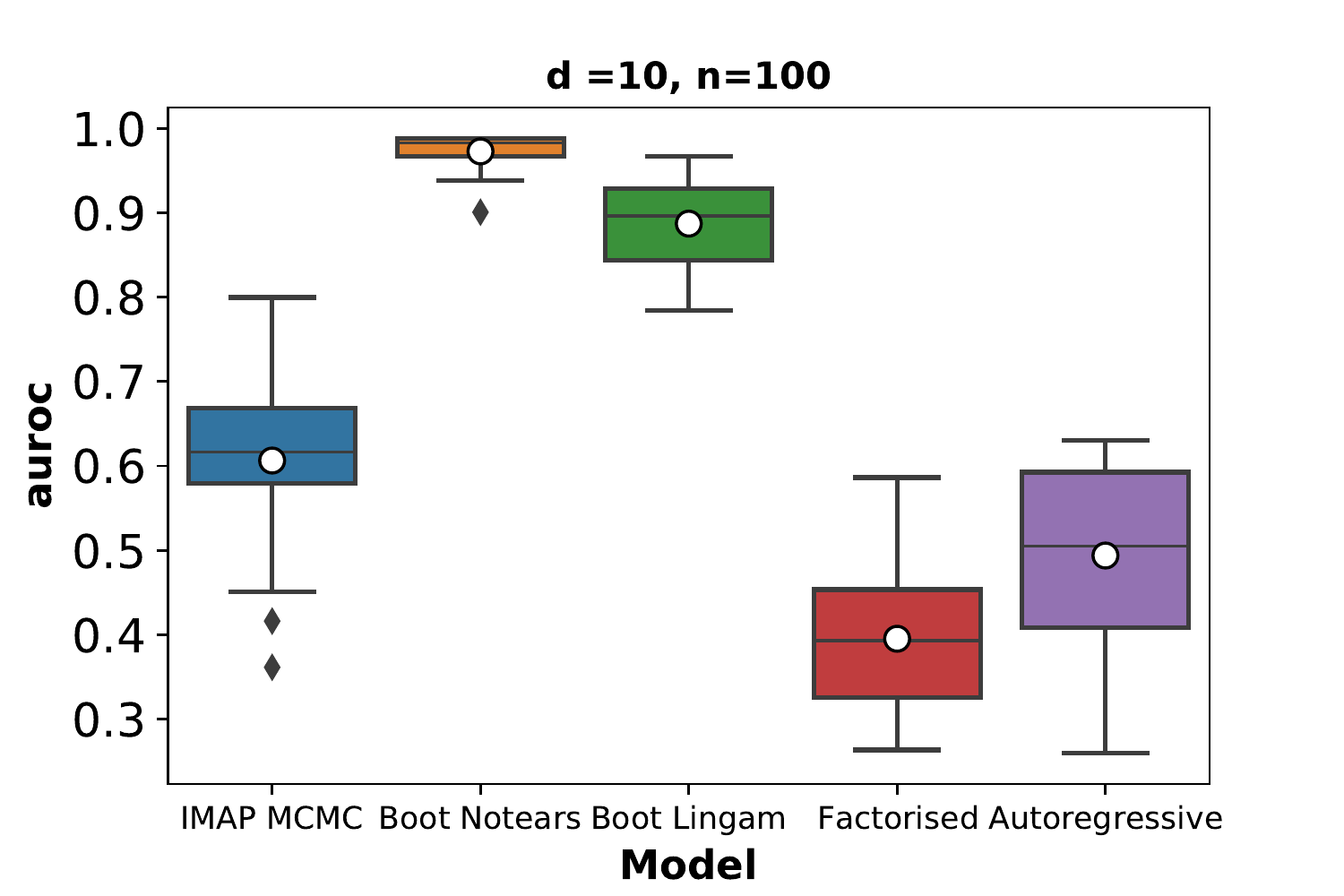}
    \end{subfigure}%
    ~ 
    \begin{subfigure}[t]{0.5\textwidth}
        \centering
        \includegraphics[height=1.9in]{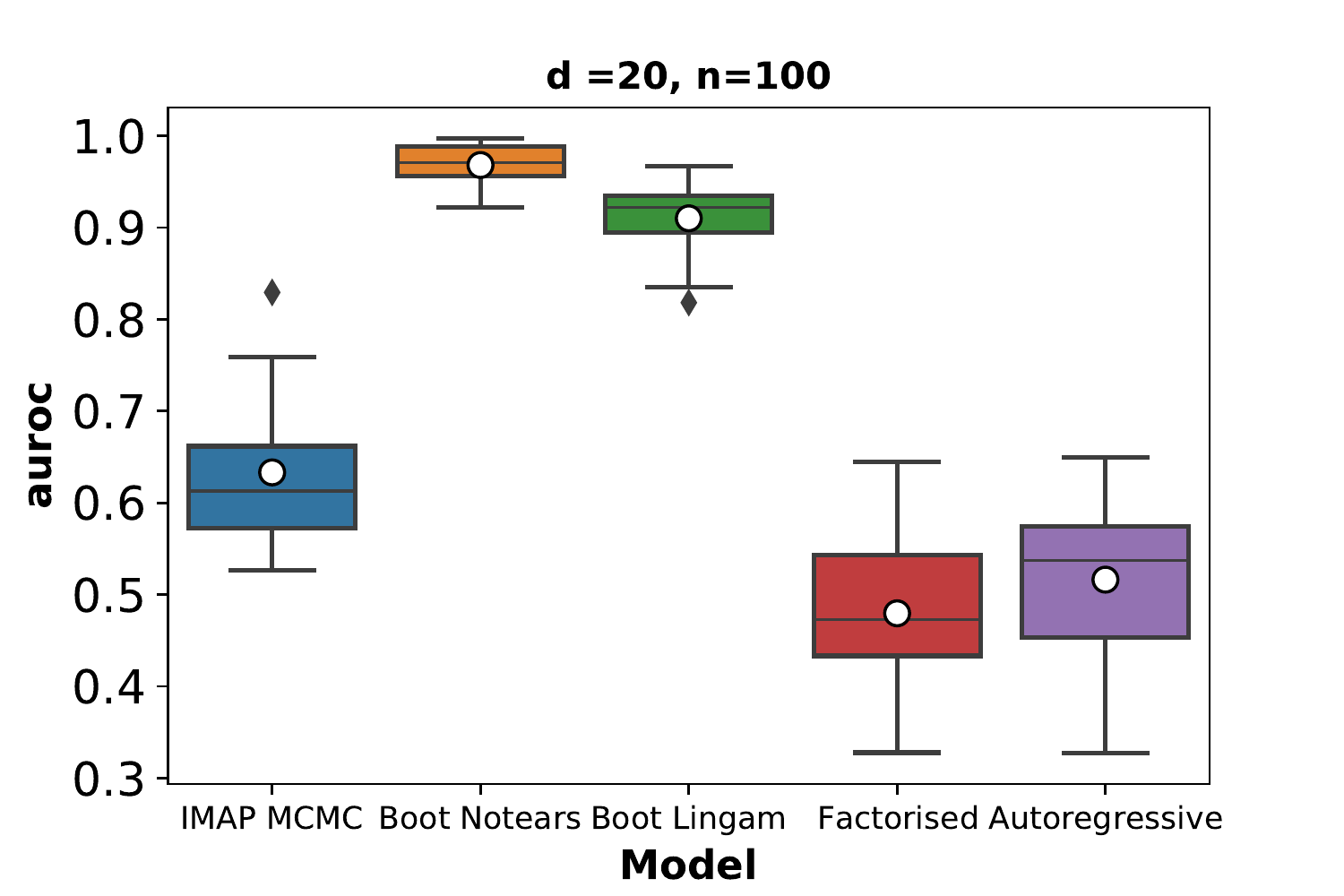}
    \end{subfigure}
    \caption{AUROC for $d=10$ and $d=20$ node ER random graphs (higher is better). Results obtained using 20 different random graphs.}
    \label{fig:auroc}
\end{figure}

 Figure~\ref{fig:exp_shd} reports the $\E[\text{SHD}]$ of the approximation for different number of nodes. It can be seen that the autoregressive distribution of VCNs gives better performance than the factorised distribution in all the settings. In addition, the autoregressive distribution gives better results as compared to all the competitive baselines when there are 10 samples, a regime common in biological applications. When the number of samples are increased to 100, the proposed approach compares favourably to the baselines which do not involve differentiable learning based techniques, while being much better than the factorised distribution. Figure~\ref{fig:auroc} presents the AUROC of these approaches. The autoregressive distribution of VCN performs better than the learning based method of factorised distribution, while comparing favourably to IMAP MCMC. We found that Boot Notears usually performs well on this metric. However, DAG Bootstrap in general has some limitations. Though DAG Bootstrap can capture multiple modes, its support is limited to the DAGs estimated using the bootstrap procedure and hence does not necessarily have full support. In addition, it can be significantly slower than VCN (Figure~\ref{fig:runtime}). Therefore, the proposed approach has a good tradeoff of runtime versus performance against the evaluated metrics and is strictly favourable in terms of purely learning based approaches.
 

\subsubsection{Real Dataset}
\begin{wraptable}[]{r}{0.45\textwidth}
\caption{Results on the Dream4 gene expression dataset in terms of AUROC.}
\label{tab:d4}
\begin{tabular}{l|l}
                              & \multicolumn{1}{c}{\textbf{AUROC}} \\ \hline\hline
IMAP MCMC                     & 0.438 $\pm$ 0.02                              \\
Boot Notears                  & 0.330                               \\
Boot Lingam                   & \textbf{0.689}                      \\
Factorised                    & 0.489 $\pm$ 0.01                               \\
\textbf{Autoregressive (VCN)} & 0.519 $\pm$ 0.03                     
\end{tabular}
\end{wraptable}
We evaluate our method in terms of AUROC on edge feature probabilities against all the baselines on the \emph{Dream4} in-silico network challenge on gene regulation. As the typical metric for this challenge is the AUROC, we restrict our focus to this metric on this dataset. In particular, we examine the multifactorial dataset
consisting of ten nodes and ten observations. Table~\ref{tab:d4} reports the AUROC of all the methods including the baselines. As the table suggests, our method achieves favourable performance while still being faster. In addition, we note that the underlying true graph corresponding to this dataset need not be a DAG, therefore introducing model misspacification.


\section{Conclusion}\label{sec:conclusion}

 We present a variational inference approach to perform Bayesian inference over the unknown causal structure in the structural causal model. Since the posterior over graph structures is multimodal in nature due to non-identifiability, we parameterise the variational family as an autoregressive distribution and model it using an LSTM. While we explore getting uncertainty estimates for linear models, it could be relevant to extend this to the case of non-linear models. One of the main limitations of this approach is the scalability of this method to larger dimensions. While a score function estimator might be limiting in higher dimensions due to higher variance, advanced variance reduction techniques could be used~\citep{grathwohl2017backpropagation}. Further improvements to the proposed method can be obtained by combining with normalising flows for discrete variables~\citep{tran2019discrete, hoogeboom2019integer}. Nevertheless, the uncertainty estimates obtained in this framework can be useful for selecting the most informative interventions, performing budgeted interventional experiments as well as representation learning of high-dimensional signals like natural images~\citep{bengio2019meta}.   
 

\clearpage

\bibliography{main}

\begin{thebibliography}{38}
\providecommand{\natexlab}[1]{#1}
\providecommand{\url}[1]{\texttt{#1}}
\expandafter\ifx\csname urlstyle\endcsname\relax
  \providecommand{\doi}[1]{doi: #1}\else
  \providecommand{\doi}{doi: \begingroup \urlstyle{rm}\Url}\fi

\bibitem[Agrawal et~al.(2018)Agrawal, Uhler, and Broderick]{agrawal2018minimal}
Raj Agrawal, Caroline Uhler, and Tamara Broderick.
\newblock Minimal i-map mcmc for scalable structure discovery in causal dag
  models.
\newblock In \emph{International Conference on Machine Learning}, pages 89--98.
  PMLR, 2018.

\bibitem[Agrawal et~al.(2019)Agrawal, Squires, Yang, Shanmugam, and
  Uhler]{agrawal2019abcd}
Raj Agrawal, Chandler Squires, Karren Yang, Karthik Shanmugam, and Caroline
  Uhler.
\newblock Abcd-strategy: Budgeted experimental design for targeted causal
  structure discovery.
\newblock \emph{arXiv preprint arXiv:1902.10347}, 2019.

\bibitem[Bengio et~al.(2019)Bengio, Deleu, Rahaman, Ke, Lachapelle, Bilaniuk,
  Goyal, and Pal]{bengio2019meta}
Yoshua Bengio, Tristan Deleu, Nasim Rahaman, Rosemary Ke, S{\'e}bastien
  Lachapelle, Olexa Bilaniuk, Anirudh Goyal, and Christopher Pal.
\newblock A meta-transfer objective for learning to disentangle causal
  mechanisms.
\newblock \emph{arXiv preprint arXiv:1901.10912}, 2019.

\bibitem[Blei et~al.(2017)Blei, Kucukelbir, and McAuliffe]{blei2017}
David~M. Blei, Alp Kucukelbir, and Jon~D. McAuliffe.
\newblock {Variational Inference: A Review for Statisticians}.
\newblock \emph{Journal of the American Statistical Association}, 112\penalty0
  (518):\penalty0 859–877, Apr 2017.
\newblock ISSN 1537-274X.
\newblock \doi{10.1080/01621459.2017.1285773}.
\newblock URL \url{http://dx.doi.org/10.1080/01621459.2017.1285773}.

\bibitem[B{\"u}hlmann et~al.(2014)B{\"u}hlmann, Peters, Ernest,
  et~al.]{buhlmann2014cam}
Peter B{\"u}hlmann, Jonas Peters, Jan Ernest, et~al.
\newblock Cam: Causal additive models, high-dimensional order search and
  penalized regression.
\newblock \emph{The Annals of Statistics}, 2014.

\bibitem[Chickering(2002)]{chickering2002optimal}
David~Maxwell Chickering.
\newblock Optimal structure identification with greedy search.
\newblock \emph{Journal of machine learning research}, 2002.

\bibitem[Eberhardt and Scheines(2007)]{eberhardt2007interventions}
Frederick Eberhardt and Richard Scheines.
\newblock Interventions and causal inference.
\newblock \emph{Philosophy of science}, 74\penalty0 (5):\penalty0 981--995,
  2007.

\bibitem[Ellis and Wong(2008)]{ellis2008learning}
Byron Ellis and Wing~Hung Wong.
\newblock Learning causal bayesian network structures from experimental data.
\newblock \emph{Journal of the American Statistical Association}, 2008.

\bibitem[Friedman and Koller(2003)]{friedman2003being}
Nir Friedman and Daphne Koller.
\newblock Being bayesian about network structure. a bayesian approach to
  structure discovery in bayesian networks.
\newblock \emph{Machine learning}, 50\penalty0 (1):\penalty0 95--125, 2003.

\bibitem[Geiger et~al.(2002)Geiger, Heckerman, et~al.]{geiger2002parameter}
Dan Geiger, David Heckerman, et~al.
\newblock Parameter priors for directed acyclic graphical models and the
  characterization of several probability distributions.
\newblock \emph{The Annals of Statistics}, 2002.

\bibitem[Grathwohl et~al.(2017)Grathwohl, Choi, Wu, Roeder, and
  Duvenaud]{grathwohl2017backpropagation}
Will Grathwohl, Dami Choi, Yuhuai Wu, Geoffrey Roeder, and David Duvenaud.
\newblock Backpropagation through the void: Optimizing control variates for
  black-box gradient estimation.
\newblock \emph{arXiv preprint arXiv:1711.00123}, 2017.

\bibitem[Grzegorczyk and Husmeier(2008)]{grzegorczyk2008improving}
Marco Grzegorczyk and Dirk Husmeier.
\newblock Improving the structure mcmc sampler for bayesian networks by
  introducing a new edge reversal move.
\newblock \emph{Machine Learning}, 2008.

\bibitem[Hauser and B{{\"u}}hlmann(2012)]{JMLR:v13:hauser12a}
Alain Hauser and Peter B{{\"u}}hlmann.
\newblock Characterization and greedy learning of interventional markov
  equivalence classes of directed acyclic graphs.
\newblock \emph{Journal of Machine Learning Research}, 2012.
\newblock URL \url{http://jmlr.org/papers/v13/hauser12a.html}.

\bibitem[Heckerman et~al.(1999)Heckerman, Meek, and
  Cooper]{heckerman1999bayesian}
David Heckerman, Christopher Meek, and Gregory Cooper.
\newblock A bayesian approach to causal discovery.
\newblock \emph{Computation, causation, and discovery}, 1999.

\bibitem[Heinze-Deml et~al.(2018{\natexlab{a}})Heinze-Deml, Maathuis, and
  Meinshausen]{heinze2018causal}
Christina Heinze-Deml, Marloes~H Maathuis, and Nicolai Meinshausen.
\newblock Causal structure learning.
\newblock \emph{Annual Review of Statistics and Its Application}, 5:\penalty0
  371--391, 2018{\natexlab{a}}.

\bibitem[Heinze-Deml et~al.(2018{\natexlab{b}})Heinze-Deml, Peters, and
  Meinshausen]{heinze2018invariant}
Christina Heinze-Deml, Jonas Peters, and Nicolai Meinshausen.
\newblock Invariant causal prediction for nonlinear models.
\newblock \emph{Journal of Causal Inference}, 2018{\natexlab{b}}.

\bibitem[Hochreiter and Schmidhuber(1997)]{hochreiter1997long}
Sepp Hochreiter and J{\"u}rgen Schmidhuber.
\newblock Long short-term memory.
\newblock \emph{Neural computation}, 1997.

\bibitem[Hoogeboom et~al.(2019)Hoogeboom, Peters, Berg, and
  Welling]{hoogeboom2019integer}
Emiel Hoogeboom, Jorn~WT Peters, Rianne van~den Berg, and Max Welling.
\newblock Integer discrete flows and lossless compression.
\newblock \emph{arXiv preprint arXiv:1905.07376}, 2019.

\bibitem[Ke et~al.(2019)Ke, Bilaniuk, Goyal, Bauer, Larochelle, Pal, and
  Bengio]{ke2019learning}
Nan~Rosemary Ke, Olexa Bilaniuk, Anirudh Goyal, Stefan Bauer, Hugo Larochelle,
  Chris Pal, and Yoshua Bengio.
\newblock Learning neural causal models from unknown interventions.
\newblock \emph{arXiv preprint arXiv:1910.01075}, 2019.

\bibitem[Ke et~al.(2020)Ke, Wang, Mitrovic, Szummer, Rezende,
  et~al.]{ke2020amortized}
Nan~Rosemary Ke, Jane Wang, Jovana Mitrovic, Martin Szummer, Danilo~J Rezende,
  et~al.
\newblock Amortized learning of neural causal representations.
\newblock \emph{arXiv preprint arXiv:2008.09301}, 2020.

\bibitem[Kingma and Ba(2014)]{kingma2014adam}
Diederik~P Kingma and Jimmy Ba.
\newblock Adam: A method for stochastic optimization.
\newblock \emph{arXiv preprint arXiv:1412.6980}, 2014.

\bibitem[Kuipers and Moffa(2017)]{kuipers2017partition}
Jack Kuipers and Giusi Moffa.
\newblock Partition mcmc for inference on acyclic digraphs.
\newblock \emph{Journal of the American Statistical Association}, 2017.

\bibitem[Kuipers et~al.(2014)Kuipers, Moffa, Heckerman,
  et~al.]{kuipers2014addendum}
Jack Kuipers, Giusi Moffa, David Heckerman, et~al.
\newblock Addendum on the scoring of gaussian directed acyclic graphical
  models.
\newblock \emph{Annals of Statistics}, 42\penalty0 (4):\penalty0 1689--1691,
  2014.

\bibitem[Lachapelle et~al.(2019)Lachapelle, Brouillard, Deleu, and
  Lacoste-Julien]{lachapelle2019gradient}
S{\'e}bastien Lachapelle, Philippe Brouillard, Tristan Deleu, and Simon
  Lacoste-Julien.
\newblock Gradient-based neural dag learning.
\newblock \emph{arXiv preprint arXiv:1906.02226}, 2019.

\bibitem[Madigan et~al.(1995)Madigan, York, and Allard]{madigan1995bayesian}
David Madigan, Jeremy York, and Denis Allard.
\newblock Bayesian graphical models for discrete data.
\newblock \emph{International Statistical Review/Revue Internationale de
  Statistique}, 1995.

\bibitem[Niinim{\"a}ki et~al.(2016)Niinim{\"a}ki, Parviainen, and
  Koivisto]{niinimaki2016structure}
Teppo Niinim{\"a}ki, Pekka Parviainen, and Mikko Koivisto.
\newblock Structure discovery in bayesian networks by sampling partial orders.
\newblock \emph{The Journal of Machine Learning Research}, 2016.

\bibitem[Pearl(2009)]{pearl2009causality}
Judea Pearl.
\newblock \emph{Causality}.
\newblock Cambridge university press, 2009.

\bibitem[Peters et~al.(2016)Peters, B{\"u}hlmann, and
  Meinshausen]{peters2016causal}
Jonas Peters, Peter B{\"u}hlmann, and Nicolai Meinshausen.
\newblock Causal inference by using invariant prediction: identification and
  confidence intervals.
\newblock \emph{Journal of the Royal Statistical Society: Series B (Statistical
  Methodology)}, 2016.

\bibitem[Peters et~al.(2017)Peters, Janzing, and
  Sch{\"o}lkopf]{peters2017elements}
Jonas Peters, Dominik Janzing, and Bernhard Sch{\"o}lkopf.
\newblock \emph{Elements of causal inference}.
\newblock The MIT Press, 2017.

\bibitem[Prill et~al.(2010)Prill, Marbach, Saez-Rodriguez, Sorger, Alexopoulos,
  Xue, Clarke, Altan-Bonnet, and Stolovitzky]{prill2010towards}
Robert~J Prill, Daniel Marbach, Julio Saez-Rodriguez, Peter~K Sorger,
  Leonidas~G Alexopoulos, Xiaowei Xue, Neil~D Clarke, Gregoire Altan-Bonnet,
  and Gustavo Stolovitzky.
\newblock Towards a rigorous assessment of systems biology models: the dream3
  challenges.
\newblock \emph{PloS one}, 2010.

\bibitem[Ramsey et~al.(2017)Ramsey, Glymour, Sanchez-Romero, and
  Glymour]{ramsey2017million}
Joseph Ramsey, Madelyn Glymour, Ruben Sanchez-Romero, and Clark Glymour.
\newblock A million variables and more: the fast greedy equivalence search
  algorithm for learning high-dimensional graphical causal models, with an
  application to functional magnetic resonance images.
\newblock \emph{International journal of data science and analytics}, 2017.

\bibitem[Schaffter et~al.(2011)Schaffter, Marbach, and
  Floreano]{schaffter2011genenetweaver}
Thomas Schaffter, Daniel Marbach, and Dario Floreano.
\newblock Genenetweaver: in silico benchmark generation and performance
  profiling of network inference methods.
\newblock \emph{Bioinformatics}, 2011.

\bibitem[Shimizu(2014)]{shimizu2014lingam}
Shohei Shimizu.
\newblock {LiNGAM: Non-Gaussian methods for estimating causal structures}.
\newblock \emph{Behaviormetrika}, 2014.

\bibitem[Spirtes et~al.(2000)Spirtes, Glymour, Scheines, and
  Heckerman]{spirtes2000causation}
Peter Spirtes, Clark~N Glymour, Richard Scheines, and David Heckerman.
\newblock \emph{Causation, prediction, and search}.
\newblock MIT press, 2000.

\bibitem[Tran et~al.(2019)Tran, Vafa, Agrawal, Dinh, and
  Poole]{tran2019discrete}
Dustin Tran, Keyon Vafa, Kumar Agrawal, Laurent Dinh, and Ben Poole.
\newblock Discrete flows: Invertible generative models of discrete data.
\newblock In \emph{Advances in Neural Information Processing Systems}, 2019.

\bibitem[Williams(1992)]{williams1992simple}
Ronald~J Williams.
\newblock Simple statistical gradient-following algorithms for connectionist
  reinforcement learning.
\newblock \emph{Machine learning}, 1992.

\bibitem[Yu et~al.(2019)Yu, Chen, Gao, and Yu]{yu2019dag}
Yue Yu, Jie Chen, Tian Gao, and Mo~Yu.
\newblock Dag-gnn: Dag structure learning with graph neural networks.
\newblock \emph{arXiv preprint arXiv:1904.10098}, 2019.

\bibitem[Zheng et~al.(2018)Zheng, Aragam, Ravikumar, and Xing]{zheng2018dags}
Xun Zheng, Bryon Aragam, Pradeep~K Ravikumar, and Eric~P Xing.
\newblock Dags with no tears: Continuous optimization for structure learning.
\newblock In \emph{Advances in Neural Information Processing Systems}, pages
  9472--9483, 2018.

\end{thebibliography}
\bibliographystyle{plainnat}

\newpage
\appendix
\section{Proof of Proposition~\ref{prop:elbo_scm}}
\begin{proof}
\begin{equation*}
    \begin{split}
        &\argmin_{\phi} \kl{\left(q_{\phi}( \G) || p(\G | \gD)\right)}\\
        &\stackrel{Eq. \ref{eq:true_posterior}}{=}\argmin_{\phi} \E_{q_\phi(\G)}[\log p(\gD|\G) + \log p(\gD) + \log q_\phi(\G)- \log p(\G)]\\[1ex]
        &=\argmin_{\phi} \E_{q_\phi(\G)}\left[-\log p(\gD|\G)\right] + \log p(\gD) + \kl\left(q_\phi(\G)||p(\G)\right) \\[1ex]
    \end{split}
\end{equation*}
which gives us a lower bound on the marginal log-likelihood of the data.
\begin{equation*}
\begin{split}
    \log p(\gD) \geq \E_{q_\phi(\G)}\left[\log p(\gD|\G)\right] - \kl\left(q_\phi(\G)||p(\G)\right) \\
\end{split}
\end{equation*}
\end{proof}

\section{Detailed Experimental Setup}
\subsection{Synthetic Data Generation}
For generating synthetic data, we follow the procedure of NOTEARS~\cite{zheng2018dags}. We sample a DAG at random from an Erdos-Renyi model with expected number of edges equal to $d$. We only select the DAG if it has atleast two graphs inside the MEC. The edge weights are sampled independently at random from a Gaussian prior with a mean of $2$ and variance of $1$.The noise variables of the SCM are Gaussian mean 0 and variance of 1. 
\subsection{Experimental Details}
Each model was then trained 20 times with different graphs (and hence datasets) for $n=10$ and $n=100$ data samples. For the plots of the true posterior comparison, we focus on $n=10$ samples. In addition, we also take the Hellinger distance instead of other divergences as the probabilities of many graph configurations are almost 0 and the log probabilities become numerically infeasible to tractably compute. For the factorised distribution and VCN, the models were trained with a learning rate of $1e-2$ with Adam optimiser~\citep{kingma2014adam} for 30k epochs where the Gibbs temperature of the DAG constraint $\lambda_t$ is annealed from 10 to 1000 with the following annealing schedule:
\begin{equation}
    \lambda_t(i) = \text{min\_temp} + 10^{\frac{-2\max(0, k-1.1i)}{k}}(\text{max\_temp} - \text{min\_temp}) 
\end{equation}
where $k$ is the total number of epochs and $i$ is the current epoch. We set $\lambda_s$ to 0.01. We also note that $Z_d$ is the partition function of the Gibbs distribution which in practice cannot be computed easily for data with dimensionality $d\geq 5$. However, we note that computing this is not required for the optimisation of the ELBO (Eq.~\ref{eq:elbo}) as it turns out to be a constant.
\subsection{BGe score}\label{app:bge}
To compute the BGe score, the parameters $\theta$ have the following prior:
\begin{equation}
    \begin{split}
        \btheta &\sim \mathcal{N}(\btheta|\mu, W^{-1})\\
        p(\mu|\gamma, W, \alpha_w) &= \mathcal{N}(\mu|\gamma, (\alpha_\mu W)^{-1})\\
        p(W|\alpha_\lambda, T) &:= \text{Wishart}(W|\alpha_\lambda, T)
    \end{split}
\end{equation}
We set $\gamma$ to 2, $\alpha_\lambda$ to 10 if $d\leq5$ else we set it to 1000. $T$ is set to $\alpha_\mu\frac{\alpha_\lambda -d - 1)}{\alpha_\mu + 1}\cdot\mathbb{I}^{d\times d}$. 
The posterior for the above equation can be obtained with the same distribution family in close form and is given by:
\begin{equation}
    \begin{split}
    p(\mu|W,\gD) &:= \mathcal{N}(\mu|\gamma',W')\\
    \gamma' &=\frac{\alpha_\mu \gamma + n \bar{x}_n}{\alpha_\mu + n}\\
    W' &= (\alpha_\mu + n)W
    \end{split}
\end{equation}
\begin{equation}
    \begin{split}
         p(W|\gD) &:= \text{Wishart}(\alpha_w', T')\\
         \alpha_w' &= \alpha_w + n\\
         T' &= T + S_n + \frac{\alpha_\mu n}{\alpha_\mu + n} (\gamma - \bar{x}_n)(\gamma - \bar{x}_n)^T
    \end{split}
\end{equation}
where $\bar{x}_n = \sum_{i=1}^{n} x^{(i)}$ is the sample mean and $S_n = \sum_{i=1}^n (x^{(i)}-\bar{x}_n)(x^{(i)}-\bar{x}_n)^T$ is the sample covariance.
\subsection{Baselines}\label{app:exp_base}
Here we review the baselines which have been used for comparison and outline the hyperparameters and training procedures which have been used.
\subsubsection{DAG Bootstrap}
For the DAG bootstrap method, we ran Lingam~\citep{shimizu2014lingam} and NOTEARS~\citep{zheng2018dags} by resampling (with replacement) 50 points at random for 1000 different times for $n=100$ and 252 times with 5 points for $n=10$. An empirical estimate of the posterior $P(\G|\X) = w_{\G,\X}\mathbbm{1}(\G\in \G_{t})$ is obtained where $\G_{t}$ is the set of all graphs predicted during the Bootstrap procedure and $w_{\G,\X}$ is the empirical normalised frequency count. 
\subsubsection{IMAP MCMC}
We follow the default hyperparameters given in the IMAP MCMC paper~\citep{agrawal2018minimal}. In particular, we run the MCMC chain for $1e5$ iterations with a burn-in period of $2e4$ iterations. We use a thinning factor of 100 and use the same parameters for BGe score calculation as that for our approach.

\subsubsection{Open Source Codes for Baselines}
We use the following open source implementations for the baselines:
\begin{itemize}
    \item NOTEARS: \url{https://github.com/xunzheng/notears} with Apache-2.0 License.
    \item Lingam: \url{https://github.com/cdt15/lingam} with MIT License.
    \item IMAP MCMC: \url{https://github.com/miraep8/Minimal_IMAP_MCMC}.
\end{itemize}
\subsection{Real Dataset}

For the Dream4~\citep{schaffter2011genenetweaver} gene-expression dataset, we use one of the gene expressions from the multifactorial dataset with ten nodes and ten observations. We also note that all the graphs in Dream4 do not necessarily correspond to DAGs, thereby introducing a crucial model misspacification in many of the models presented. We hypothesize that the inferior results of many of the techniques including the proposed approach suffer from this because of the non-DAGness.

\end{document}